\documentclass{article}

\usepackage[preprint]{neurips_2026}
\makeatletter
\def\@trackname{}
\makeatother
\usepackage[utf8]{inputenc}
\usepackage[T1]{fontenc}
\usepackage{hyperref}
\usepackage{url}
\usepackage{booktabs}
\usepackage{amsmath,amssymb}
\usepackage{microtype}
\usepackage{xspace}
\usepackage{xcolor}
\usepackage{colortbl}
\usepackage{graphicx}
\usepackage{fontawesome5}
\usepackage{float}
\definecolor{linkc}{RGB}{31,78,121}   
\definecolor{urlc}{RGB}{21,118,138}   
\hypersetup{
  colorlinks=true,
  linkcolor=linkc,
  citecolor=linkc,
  urlcolor=urlc,
  filecolor=linkc,
  anchorcolor=linkc,
  pdfborder={0 0 0}
}
\usepackage{sectsty}
\sectionfont{\color{linkc}}
\subsectionfont{\color{linkc}}
\subsubsectionfont{\color{linkc}}
\paragraphfont{\color{linkc}}
\usepackage{tcolorbox}
\tcbuselibrary{skins,breakable}
\setcitestyle{authoryear,round}
\usepackage{fancyhdr}
\definecolor{skyblue}{RGB}{135,206,235}
\newtcolorbox{neuripsabstract}[1][]{%
  enhanced,
  colback=skyblue!14!white,
  colframe=skyblue!55!white,
  arc=6pt,
  boxrule=0.5pt,
  leftrule=2.5pt,
  left=12pt,
  right=12pt,
  top=10pt,
  bottom=10pt,
  fontupper=\small,
  center title,
  title=Abstract,
  #1
}

\newcommand{\method}{\textsc{VideoHarness-RSI}\xspace}
\newcommand{\uniformh}{\textsc{Uniform-40}\xspace}
\newcommand{\captionnav}{\textsc{CaptionNavigate}\xspace}
\newcommand{\weakft}{\textsc{StatedTimeAddressDecode}\xspace}
\newcommand{\cardinalityledger}{\textsc{CardinalityLedger}\xspace}
\newcommand{\clipknn}{\textsc{CLIP-kNN}\xspace}
\newcommand{\captionknn}{\textsc{Caption-kNN}\xspace}
\newcommand{\worldmmstyle}{\textsc{WorldMM-style}\xspace}
\newcommand{\worldmmmax}{\textsc{WorldMM-style-MaxBudget}\xspace}
\newcommand{\homerstyle}{\textsc{Homer-style}\xspace}
\newcommand{\videoseal}{\textsc{VideoSeal-matched}\xspace}
\newcommand{\geminiflash}{\textsc{Gemini-2.5-Flash}\xspace}
\newcommand{\gptfive}{\textsc{GPT-5}\xspace}
\newcommand{\aksh}{\textsc{AKS}\xspace}
\newcommand{\prfh}{\textsc{AdaptiveDensity-PRF}\xspace}

\title{VideoHarness-RSI: Recursive Harness Self-Improvement for Long-Video Understanding with Frozen Vision-Language Models}

\author{
Guoyang Xu \\
Tencent \\
\texttt{nikolaxu@tencent.com} \\
\and
Hao Chen \\
Tencent \\
\texttt{agihaochen@tencent.com} \\
}

\begin{document}

\pagestyle{plain}
\fancyhead[L]{%
  \includegraphics[height=1.1cm]{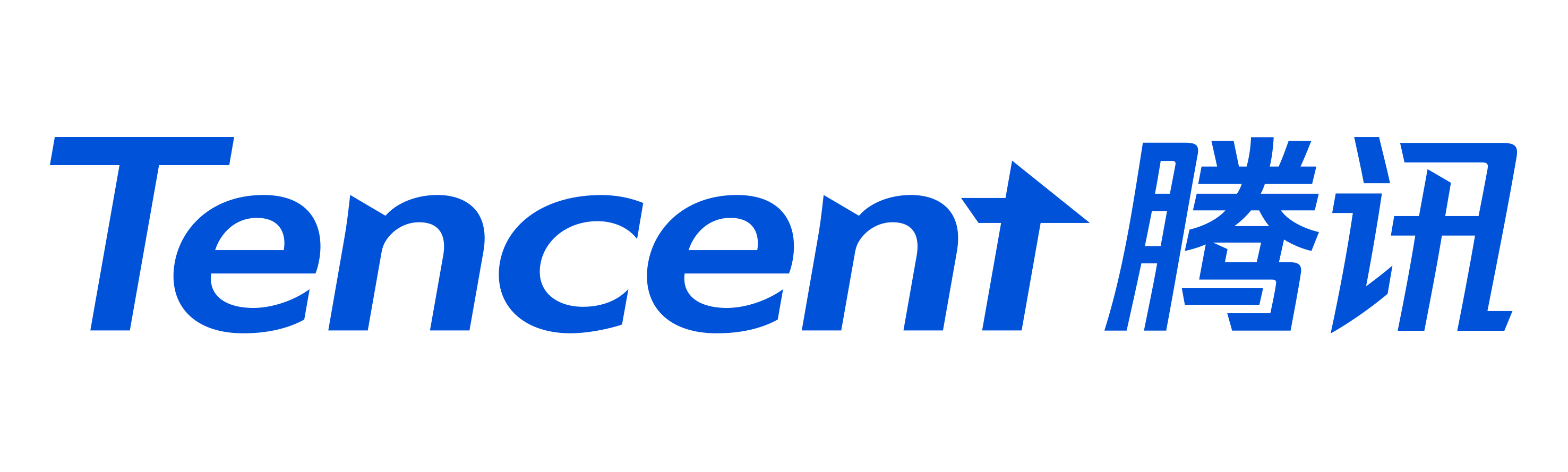}%
}
\fancyhead[C]{}
\fancyhead[R]{}
\fancyfoot[C]{}
\renewcommand{\headrulewidth}{0.4pt}
\renewcommand{\footrulewidth}{0pt}

\makeatletter
\renewcommand{\@maketitle}{%
  \newpage
  \null
  \vskip 2em%
  \begin{center}%
    {\LARGE \@title \par}%
    \vskip 1.5em%
    {\large
      \lineskip .5em%
      \begin{tabular}[t]{c}
        \@author
      \end{tabular}\par}%
  \end{center}%
  \par\vskip 1.5em%
}
\makeatother

\maketitle
\thispagestyle{fancy}

\begin{neuripsabstract}

Long-video understanding depends not only on the capability of a vision-language model (VLM), but also on how its limited context is constructed from a much longer video. Existing systems typically introduce hand-designed sampling, retrieval, memory, or agentic control strategies, making the context-construction program itself difficult to study as an independent optimization target. We introduce \method, a controlled framework that recursively searches executable context constructors around a frozen VLM while keeping the answering model and interface fixed. We study this baseline under complementary weak- and strong-initialization regimes. From a weak uniform constructor, recursive search progressively discovers more structured context-construction programs; from a stronger AKS harness, the same process further advances an already competitive hand-crafted frontier. The resulting harness retains its advantage under a matched cumulative visual-token control and transfers directly to additional long-video benchmarks without further search. Together, these results establish executable context construction as a distinct optimization layer and provide an auditable baseline for studying harness discovery, transfer, and efficiency around frozen VLMs.

\end{neuripsabstract}

\begin{center}
    \small
    \href{https://github.com/Tencent/VideoHarness-RSI}{\faGithub\ \texttt{https://github.com/Tencent/VideoHarness-RSI}}
\end{center}
\vspace{-4pt}

\section{Introduction}

Long-form video remains difficult for vision-language models (VLMs) even when the underlying model is strong. Relevant evidence may be sparse, temporally distant, and surrounded by thousands of irrelevant observations. Exhaustively presenting a long video is therefore impractical: performance depends not only on what a VLM can infer, but also on the context that its surrounding system chooses to expose~\citep{wang2025lvbench,shen2024longvu}. Long-video understanding is partly a context-construction problem.

Existing systems address this bottleneck through compression, retrieval, external memory, temporal navigation, and iterative evidence acquisition~\citep{shen2024longvu,wang2024videoagent,zhang2025dvd}. Yet they often change several components at once---evidence representation, retrieval, tools, reasoning workflow, or even the model---making it difficult to isolate how much improvement can come from the executable context-construction program alone.

We ask whether such a harness can improve recursively while the underlying VLM and its interface remain fixed. \method treats the harness as a program that organizes, retrieves, and packs evidence before a frozen VLM answers. An outer loop proposes executable mutations, evaluates them end to end, and retains empirically better programs as the next search frontier. This is a controlled instance of automated harness design: the proposer searches executable programs that manage what a fixed downstream model observes. Here, recursive self-improvement means repeated improvement of harness code through proposal and task evaluation, not improvement of the VLM's parameters or intrinsic intelligence.

\begin{figure*}[t]
\centering
\includegraphics[width=1\textwidth]{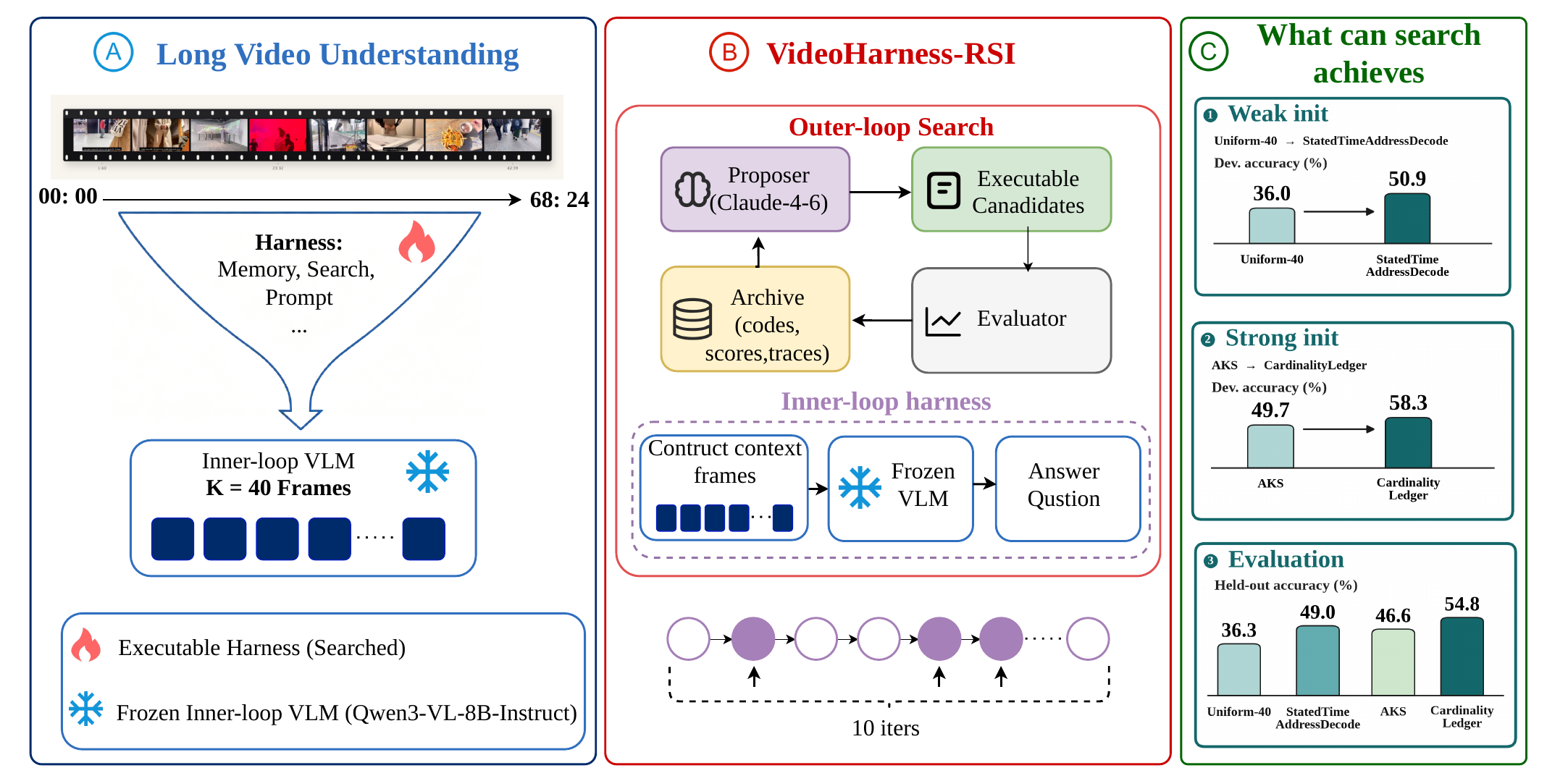}
\caption{Overview of the controlled harness-search setting. The outer loop modifies executable context-construction code around a frozen VLM, while the answering model and decoding configuration remain fixed. Development data are used only for candidate evaluation and frontier promotion. The selected harness is then frozen for held-out and cross-benchmark evaluation. In the main RSI protocol, every VLM call contains at most $K=40$ visual frames.}
\label{fig:overview}
\end{figure*}

This controlled formulation separates model capability from context-construction capability and makes program-level search itself observable. We study the baseline under two initialization regimes. Uniform initialization exposes how recursive search builds increasingly structured context construction from a weak starting point, while our main AKS initialization tests whether the same process can further advance a strong hand-crafted harness. Figure~\ref{fig:overview} summarizes the setting. Our contributions are threefold:
\begin{itemize}
    \item We formalize long-video context construction as an executable harness-optimization problem around a frozen VLM and fixed answering interface.
    \item We provide \method, a propose--execute--evaluate--retain baseline that recursively searches the program controlling evidence acquisition and context construction.
    \item We provide an auditable evaluation protocol built around complete search lineage, per-question outputs, execution traces, held-out evaluation, and context-cost accounting, enabling searched harnesses to be studied as programs rather than only as final scores.
\end{itemize}

\vspace{-2pt}
\section{Related Work}

\subsection{Long-Video Context Construction}
Long-video systems must represent more evidence than a VLM can consume directly. Existing approaches use adaptive compression, frame selection, caption or embedding indexes, and retrieval to expose a smaller question-relevant context~\citep{shen2024longvu,wang2025lvbench,fu2024videomme,zhou2025mlvu}. These works design or learn particular mechanisms; our question is whether the executable context-construction mechanism itself can be searched recursively.

\subsection{Agentic Video Understanding}
VideoAgent and Deep Video Discovery use agents to actively acquire question-relevant evidence~\citep{wang2024videoagent,zhang2025dvd}, while WorldMM, Homer, and VideoSEAL explore multimodal memory, hierarchical reasoning, and planner--inspector control~\citep{vhYeo2026WorldMM,vhJi2026Homer,vhQiu2026VideoSEAL}. MetaVideoAgent studies diagnosis-guided evolution of modular video-agent pipelines~\citep{cui2026metavideoagent}. Our focus is narrower: we isolate the executable program that constructs the final context and make that program the object of outer-loop search while keeping the answering model fixed.

\subsection{Automated Agent and Harness Optimization}
ADAS and AFlow optimize agent designs or workflows from execution feedback~\citep{hu2025adas,zhang2025aflow}; Meta-Harness directly searches model-harness code from prior source, scores, and traces~\citep{lee2026metaharness}, with related program-level harness synthesis in other domains~\citep{liu2026agentflow}. These efforts connect to program synthesis and LLM-guided program search~\citep{gulwani2017programsynthesis,koza1992genetic,ellis2021dreamcoder,romeraparedes2024funsearch}. \method does not propose a new general-purpose search algorithm; it provides a controlled long-video instantiation in which the mutable object is restricted to executable context construction.

\section{VideoHarness-RSI}

\subsection{Executable Context Constructors}
Let $\mathcal{D}_{\mathrm{dev}}=\{(V_i,q_i,y_i)\}$ be a development set, $M$ a frozen VLM, and $K$ a constraint on the final visual context. A harness $H\in\mathcal{H}_{\mathrm{exec}}$ is an executable program that maps a video--question pair to bounded multimodal context,
\begin{equation}
    C_H = H(V,q;K), \qquad \hat{y}=M(C_H,q).
    \label{eq:harness}
\end{equation}
We optimize development accuracy over executable context constructors,
\begin{equation}
    H^* = \arg\max_{H\in\mathcal{H}_{\mathrm{exec}}}
    \frac{1}{|\mathcal{D}_{\mathrm{dev}}|}
    \sum_i \mathbf{1}\big[M(H(V_i,q_i;K),q_i)=y_i\big].
    \label{eq:objective}
\end{equation}
The parameters and decoding configuration of $M$ remain fixed, and every VLM invocation in a controlled comparison respects the same $K$. A multi-stage harness may make several bounded calls, so cumulative answer-time frames can exceed $K$ even though the context visible in any single call remains fixed. The controlled variable is therefore executable context construction rather than model training or per-call visual capacity.

For analysis, a harness can be decomposed as
\begin{equation}
H(V,q;K)=\operatorname{Pack}_H\!\left(
\operatorname{Read}_H(\operatorname{Write}_H(V),q),K\right).
\label{eq:wrp}
\end{equation}
\textsc{Write} constructs an addressable representation such as a video stream, caption list, or embedding index; \textsc{Read} retrieves question-conditioned evidence; and \textsc{Pack} orders and formats the bounded context supplied to $M$. This is an analytical decomposition of one context constructor, not three separately optimized objectives. A harness may edit, replace, or compose behavior across this path; no particular memory, sampler, or retriever is mandatory.

\subsection{Recursive Harness Search}
At generation $t$, let $F_t$ denote the incumbent accuracy frontier and $\mathcal{A}_t$ an archive containing historical harness code, scores, and evaluation traces. The outer-loop proposer acts as the automated designer: it does not directly perform the downstream video-understanding task, but proposes changes to the harness that determines what the frozen VLM will observe. It generates executable candidates
\begin{equation}
    \{H_{t,j}\}_{j=1}^{m_t} \sim P(F_t,\mathcal{A}_t).
    \label{eq:proposal}
\end{equation}
The proposer is not assigned a fixed module to edit or a mandatory failure-attribution procedure: choosing the mutation direction is part of the proposal. Each candidate is smoke-tested and evaluated end to end on $\mathcal{D}_{\mathrm{dev}}$. With $S(H)$ denoting development accuracy, the update rule is deliberately strict:
\begin{equation}
F_{t+1}=
\begin{cases}
H_{t,j^*}, & \text{if } S(H_{t,j^*})>S(F_t),\\
F_t, & \text{otherwise,}
\end{cases}
\qquad
j^*=\arg\max_j S(H_{t,j}).
\label{eq:update}
\end{equation}
The held-out set never enters proposal, selection, promotion, early stopping, or rollback. Promotion uses a deterministic point-estimate comparison on the development split; a strict increase is not a statistical guarantee of generalization. The search frontier and code lineage need not be identical because a proposer may reuse or compose archived code.

\subsection{Search Interface and Constraints}
The mutable object is executable context-construction code. A candidate may change evidence representation, retrieval, navigation, selection, packing, auxiliary prompting, or their composition. The inner VLM, answer interface, task metric, and data available to the evaluator remain fixed within a search protocol. This contract separates a model improvement from a harness improvement while leaving the program space expressive enough to include uniform access, retrieval, and multi-stage navigation.

\section{Experimental Setup}

\subsection{Datasets and Evaluation Splits}
\paragraph{LVBench.}
LVBench public metadata contains 103 videos and 1,549 question--answer pairs~\citep{wang2025lvbench}. Its source videos are retrieved from YouTube. At evaluation time, 20 source videos had been removed or made inaccessible, primarily because of copyright-related availability restrictions, so our local collection contains 83 videos corresponding to 1,232 QA pairs. The unavailable videos account for the remaining 317 questions and are excluded as whole videos rather than through random question subsampling. We apply \texttt{Random(42).shuffle} once to the 1,232 examples, use positions $[0{:}350]$ as the development set for harness search, and reserve positions $[350{:}]$ (882 examples) for final held-out evaluation. The held-out questions are never used by the proposer or promotion rule.

\paragraph{Cross-benchmark evaluation.}
We evaluate direct cross-benchmark reuse on the full Video-MME (2,700 QA)~\citep{fu2024videomme} and MLVU (2,174 QA)~\citep{zhou2025mlvu} evaluation sets. The LVBench-selected \cardinalityledger harness is frozen before target-benchmark evaluation and receives no target-specific search or adaptation. We compare it with the same frozen AKS parent and uniform sampling under the target-benchmark protocol.

\subsection{Frozen Model and Context Protocol}
The main experiments freeze Qwen3-VL-8B-Instruct~\citep{bai2025qwen3vl}. Inner-loop decoding uses temperature 0 with thinking disabled. Every VLM invocation contains at most $K=40$ visual observations. A 2-fps video stream is available to harness code, and CLIP ViT-B/32 is the frozen image--text scorer when semantic image retrieval is used. The final answer interface and letter-only evaluation protocol are shared across controlled comparisons. We additionally report cumulative frames and visual tokens across answer-time calls, which may exceed 40 for multi-stage harnesses without changing the frozen per-call context.

\subsection{Search Protocol}
Candidates are generated by a Claude Opus 4.6 proposer through Claude Code. The proposer can inspect the current frontier and archive artifacts containing prior executable code, scores, and traces. We study two initialization regimes: a weak-seed trajectory from \uniformh and a main strong-seed trajectory from \aksh. Both are presented under the same sequential propose--evaluate--retain abstraction, and the development frontier is replaced only after a strict point-estimate accuracy improvement. Each reported trajectory contains ten iterations. The uniform-seeded trajectory selects \weakft, while the AKS-seeded trajectory selects \cardinalityledger. The former probes structure discovery from a minimal constructor, whereas the latter tests frontier advancement from a strong hand-crafted starting point. The Qwen and CLIP services remain fixed throughout. Alternative-capacity studies are reported separately in Appendix~\ref{app:complementary}.

\subsection{Controlled Reference Constructors}
We compare against four prespecified $K=40$ context constructors under the same VLM and answer protocol. \uniformh selects $K$ observations uniformly at answer time. \clipknn writes a 320-frame image-embedding index and maps the question directly to image neighbors. \captionknn retrieves caption neighbors and maps the hits back to their frames. \aksh balances question relevance with temporal coverage through adaptive keyframe selection~\citep{vhTang2025AKS}. These constructors provide controlled reference points under the common frozen-VLM interface.

\subsection{Evaluation Metrics}
Accuracy is the primary metric. For paired predictions on the same questions, we report exact two-sided McNemar tests and the two discordant counts. Mechanism analysis uses LVBench's annotated evidence intervals only after inference: an example is \emph{evidence-visible} when at least one selected frame lies inside an annotated window. We also report conditional accuracy, mean in-window frames, and answer-parse failures. The harness never receives these evidence annotations.

To separate search quality from context consumption, we additionally report cumulative answer-time frames and input-side context cost. For a harness $H$, the logged token cost is
\begin{equation}
    c(H)=T_{\mathrm{visual}}(H)+T_{\mathrm{text}}(H),
    \label{eq:cost}
\end{equation}
measured as average tokens per question over answer-time calls. This reporting quantity does not affect frontier promotion. Equal per-call visual capacity therefore does not imply equal cumulative computation. Full accounting boundaries are given in Appendix~\ref{app:archive}.

\section{Results}

\subsection{Main Results under Two Initialization Regimes}
Table~\ref{tab:main} reports fixed baselines together with the endpoints selected under both initialization regimes. From the weak uniform seed, recursive search selects \weakft and retains its improvement on held-out questions. From the stronger AKS seed, the same search process advances the hand-crafted frontier to \cardinalityledger, which remains the strongest selected endpoint.

\begin{table*}[t]
\centering
\caption{LVBench results. Except for the explicitly marked proprietary-VLM rows, all methods use the frozen Qwen3-VL-8B-Instruct answerer and at most $K=40$ visual observations per VLM invocation. ``Avg. frames'' and token estimates are cumulative across answer-time calls per question. The RSI-selected rows are the endpoints of the uniform- and AKS-seeded trajectories. Literature-inspired rows map published design patterns into the common interface.}
\label{tab:main}
\footnotesize
\setlength{\tabcolsep}{5pt}
\resizebox{\textwidth}{!}{%
\begin{tabular}{@{}lrrrr@{}}
\toprule
Method & Dev. acc. & Held-out acc.  & Avg. frames & Avg. input tokens \\
\midrule
\multicolumn{5}{@{}l}{\emph{Prespecified baselines}} \\
\addlinespace[1pt]
\uniformh                                      & 36.3 & 36.3 & 40.0 & 10,407 \\
\clipknn                                       & 38.6 & 41.7 & 40.0 & 10,407 \\
\captionknn                                    & 38.6 & 34.9 & 40.0 & 9,328 \\
\midrule
\multicolumn{5}{@{}l}{\emph{Hand-crafted harness baselines (literature-inspired)}} \\
\aksh~\citep{vhTang2025AKS}                    & 49.7 & 46.6 & 40.0 & 8,788 \\
\addlinespace[1pt]
\worldmmstyle~\citep{vhYeo2026WorldMM}         & ---  & 36.0 &  7.3 & $\sim 4{,}200$ \\
\worldmmmax~\citep{vhYeo2026WorldMM}           & ---  & 40.6 & 40.0 & $\sim 12{,}623$ \\
\homerstyle~\citep{vhJi2026Homer}              & ---  & 39.4 & 37.2 & 9,668 \\
\videoseal~\citep{vhQiu2026VideoSEAL}          & ---  & 37.0 & 63.4 & 8,100 \\
\midrule
\multicolumn{5}{@{}l}{\emph{Uniform sampling with proprietary VLMs}} \\
\addlinespace[1pt]
\uniformh (\geminiflash)~\citep{vhGeminiTeam2025Gemini25}
                                                  & ---  & 33.7 & 40.0 & $\approx 10{,}407^{\dagger}$ \\
\uniformh (\gptfive)~\citep{vhOpenAI2025GPT5}    & ---  & 46.5 & 40.0 & $\approx 10{,}407^{\dagger}$ \\
\midrule
\multicolumn{5}{@{}l}{\emph{RSI-selected harnesses}} \\
\addlinespace[1pt]
\weakft\ (Uniform-seeded)                    & 50.9 & 49.0 (432) & 40.0 & 9,962 \\
\rowcolor{skyblue!8}
\textbf{\cardinalityledger\ (AKS-seeded)}    & \textbf{58.3 (204)} & \textbf{54.8 (483)} & 89.5 & 19,210 \\
\bottomrule
\end{tabular}
}

\par\smallskip
{\footnotesize\raggedright
\textit{Note.} A dash indicates an unreported evaluation. Input tokens combine visual and textual answer-time context and exclude offline indexing, proposer search, output tokens, and provider pricing. The matched-token control in Table~\ref{tab:matched-token} reports visual tokens only. $^{\dagger}$ Proprietary-VLM costs are estimates based on the Uniform-40 input geometry. Accounting and re-evaluation details appear in Appendix~\ref{app:repro}.
\par
}
\end{table*}

The uniform-seeded endpoint remains above its starting reference on held-out questions, while the held-out gain of \cardinalityledger over AKS is supported by both a paired bootstrap confidence interval and an exact two-sided McNemar test (Appendix~\ref{app:repro}). Together, the two regimes show structure discovery from a weak seed and further frontier advancement from a strong hand-crafted seed.

The input-cost estimates expose a complementary trade-off. \cardinalityledger uses more cumulative answer-time context than AKS while every individual VLM call remains capped at $K=40$. Section~\ref{sec:matched-token} therefore introduces a higher-capacity AKS-90 control at approximately the same cumulative visual-token cost. The reported token counts should be read as context-volume estimates rather than end-to-end latency or monetary cost, since offline indexing and auxiliary model calls differ across harnesses.

\subsection{Matched Cumulative Visual-Token Control}
\label{sec:matched-token}
The main protocol fixes the per-call visual capacity at $K=40$, but \cardinalityledger can make multiple bounded VLM calls and therefore consumes more cumulative visual tokens than the single-call AKS baseline. We introduce \textbf{AKS-90} as a post-hoc cost control to test whether this additional visual-token consumption alone explains the gain. AKS-90 keeps the AKS selection rule unchanged but supplies 90 selected frames directly to the answering VLM in a single call. In contrast, \cardinalityledger retains the main protocol's per-call limit of at most 40 visual frames and reaches a comparable cumulative visual-token cost through its multi-stage execution.

Table~\ref{tab:matched-token} compares the searched harness with a higher-capacity AKS control at closely matched cumulative visual-token cost.

\begin{table}[t]
\centering
\caption{Matched cumulative visual-token control on LVBench. AKS-90 sends 90 frames directly to the answering VLM in one call. \cardinalityledger instead keeps every VLM call at $\leq40$ visual frames and uses multiple bounded calls. Visual-token counts are held-out averages.}
\label{tab:matched-token}
\small
\begin{tabular}{lrrrr}
\toprule
Method & Visual frames per VLM call & Avg. visual tokens & Dev.350 & Held-out 882 \\
\midrule
AKS-40 & 40 & 8,703 & 49.7 & 46.6 (411/882) \\
AKS-90 & 90 & 19,583 & 53.1 & 49.2 (434/882) \\
\rowcolor{skyblue!8}
\cardinalityledger & $\leq40$ & 19,210 & \textbf{58.3} & \textbf{54.8 (483/882)} \\
\bottomrule
\end{tabular}
\end{table}

Increasing the parent's answer-frame capacity improves accuracy, confirming that additional visual context contributes to performance. Yet the higher-capacity parent does not close the gap to the searched harness at similar cumulative visual-token cost. This pattern is consistent with gains from both context volume and the query-adaptive allocation of evidence across survey, refinement, packing, and optional ledger operations.

\subsection{Cross-Benchmark Generalization}
\label{sec:cross-benchmark}
We next test whether the LVBench-selected program transfers beyond the benchmark used for search. Table~\ref{tab:transfer} compares \cardinalityledger with the same AKS parent and uniform sampling, with all methods frozen before target-benchmark evaluation and no target-specific search or adaptation.

\begin{table}[t]
\centering
\caption{Cross-benchmark direct reuse of the LVBench-selected \cardinalityledger harness. All methods use the same frozen VLM and answering interface. \cardinalityledger preserves the per-call limit of at most 40 visual frames while allowing multi-stage execution. No target-specific search or adaptation is used.}
\label{tab:transfer}
\small
\begin{tabular}{lrrrr}
\toprule
Benchmark & \#QA & Uniform & \aksh & \cardinalityledger \\
\midrule
Video-MME~\citep{fu2024videomme} & 2,700 & 59.9 & 66.9 & \textbf{67.5} \\
MLVU~\citep{zhou2025mlvu} & 2,174 & 63.2 & 72.0 & \textbf{75.0} \\
\bottomrule
\end{tabular}
\end{table}

\cardinalityledger retains an advantage over the strong AKS parent on both MLVU and Video-MME under direct reuse, with a larger improvement on MLVU. The consistency of this ordering across the two benchmarks suggests that the program structure discovered on LVBench transfers beyond the development distribution used for search. Together with the held-out LVBench result, these evaluations show that recursive harness search can produce context constructors whose benefits extend beyond the questions used to select them.

\subsection{Literature-Inspired Harness Controls}
The literature-inspired rows in Table~\ref{tab:main} map three published design patterns into the common frozen-VLM interface: multimodal-memory routing, hierarchical memory with verification, and planner--inspector separation~\citep{vhYeo2026WorldMM,vhJi2026Homer,vhQiu2026VideoSEAL}. Their behavior under the same answerer and evaluation protocol provides a structural comparison with recursive harness search. The proprietary-VLM rows separately vary the answerer while keeping uniform sampling fixed. Implementation details appear in Appendix~\ref{app:literature-harnesses}.

\subsection{What Does Recursive Search Discover?}
\begin{figure}[t]
\centering
\includegraphics[width=\linewidth]{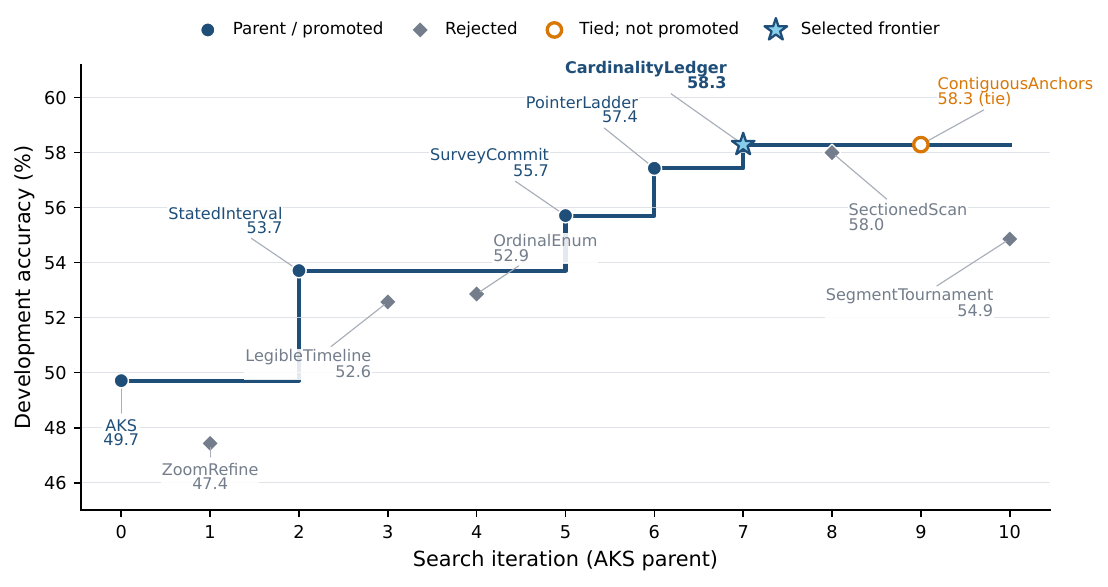}
\caption{Complete ten-candidate AKS-seeded VideoHarness-RSI trajectory. The parent is replaced only by a strict development-accuracy improvement, and held-out results are never used for promotion.}
\label{fig:search-frontier}
\end{figure}

\begin{figure}[t]
\centering
\includegraphics[width=\linewidth]{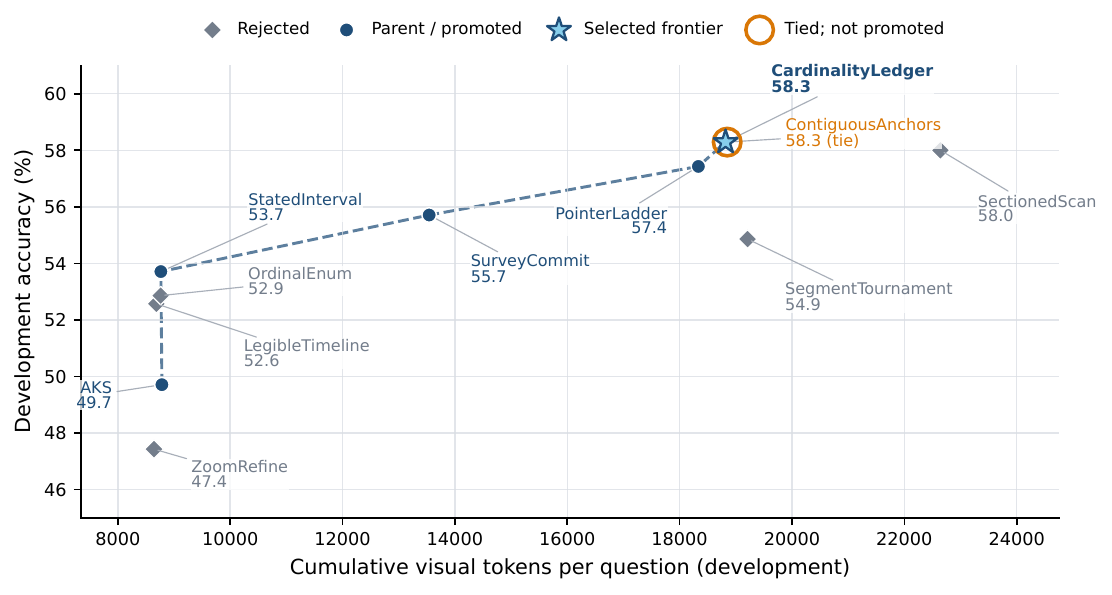}
\caption{Development accuracy versus cumulative visual tokens per question for the complete search. Every VLM invocation remains capped at $K=40$ observations; the horizontal cost reflects multiple bounded calls when present.}
\label{fig:search-pareto}
\end{figure}

Figures~\ref{fig:search-frontier} and~\ref{fig:search-pareto} show the complete AKS-seeded trajectory and its accuracy--cost trade-off. Four strict frontier updates progressively introduce explicit interval handling, survey-based evidence acquisition, pointer refinement, and cardinality-aware bookkeeping, culminating in \cardinalityledger. Subsequent candidates do not further advance the frontier. The complementary uniform-seeded trajectory in Appendix~\ref{app:archive} shows the same search process building increasingly structured context construction from a weak starting point and ultimately selecting \weakft; its improvement also persists on held-out questions.

\vspace{-2pt}
\vspace{-2pt}
Together, the frontier and Pareto views separate strict recursive improvement from the cumulative cost of constructing context. Exact candidate scores, source code, parentage, hypotheses, and traces appear in Appendix~\ref{app:archive}.

\section{Analysis}
\subsection{What Program Did RSI Discover?}
Execution traces show that \cardinalityledger is query-adaptive rather than a fixed frame selector: it can bypass search for explicit-time questions, use coarse-to-fine refinement for general questions, and add a ledger step for counting queries. Detailed execution paths and representative examples are provided in Appendix~\ref{app:cardinality-execution}.

\section{Discussion}

\paragraph{Harnesses form a distinct optimization layer.}
The studied object is the executable mapping from a long video and question to the context seen by a frozen VLM. Treating this mapping as mutable makes it possible to improve system behavior without changing model parameters and to separate harness design from model training.

\paragraph{Search can discover control flow, not only better sampling.}
The searched programs increasingly compose conditional routing, coarse-to-fine evidence acquisition, context packing, and auxiliary bookkeeping. This suggests that the useful search space is broader than frame scoring: an inference harness can evolve into a query-dependent control program that decides how much evidence acquisition and verification a question requires.

\paragraph{Capability and execution cost should be separated.}
A fixed per-call visual limit does not imply a fixed total inference budget when a harness can make multiple bounded reads. The AKS-90 control shows that increasing visual context improves the parent, but does not close the gap to \cardinalityledger at a similar cumulative visual-token cost. Accuracy and cumulative context cost should therefore be reported jointly rather than treating equal final context size as equal computation.

\paragraph{Search evidence and evaluation evidence serve different roles.}
Code, traces, and development outcomes guide program search, whereas held-out and cross-benchmark evaluations assess the selected endpoint after it is frozen. Keeping these roles separate is necessary for interpreting recursive search without turning development feedback into evidence of generalization. The MLVU and Video-MME results further show that a program selected on one benchmark can retain an advantage over its strong parent under direct reuse, while the magnitude of that advantage can vary across benchmarks.

\paragraph{Scope and limitations.}
The experiments use one frozen VLM, one data seed, and a limited number of search trajectories, so they do not establish cross-model behavior or search variance. LVBench uses the locally available subset and a question-level rather than video-disjoint split. The proprietary proposer may contain benchmark-level prior knowledge, and equal per-call visual budgets do not imply equal cumulative cost. The studied programs also build per-video contexts rather than mutable cross-question memory. Accordingly, ``recursive self-improvement'' refers to iterative improvement of executable harness programs, not recursive amplification of the VLM's underlying intelligence.

\section{Conclusion}
We introduced \method as a controlled setting for recursively optimizing executable context constructors around a frozen long-video VLM. By making the context-construction program itself the search object, the framework separates harness improvement from model training and exposes the evolution of inference-time evidence acquisition as an auditable process. Held-out and direct cross-benchmark evaluation show that the selected program can retain its advantage beyond the development questions used for search. The resulting baseline supports systematic study of program discovery, generalization, and efficiency, and provides a foundation for future work on more capable and cost-aware self-improving inference harnesses.

\clearpage
\bibliography{videoharness_refs_v8}
\clearpage
\appendix
\begin{center}
    {\LARGE\bfseries Appendix}
\end{center}
\vspace{0.25em}
\noindent\small
This appendix provides the context-constructor taxonomy, complete search trajectories, cost accounting, complementary search protocols, literature-inspired controls, and reproducibility audit.
\normalsize
\vspace{0.75em}

\section{Context-Constructor Taxonomy}
\label{app:taxonomy}
Table~\ref{tab:taxonomy} records the mechanisms using the analytical decomposition in Equation~\ref{eq:wrp}. Groups denote search parent and budget protocol, not a common ranking. In particular, the caption-kNN path maps caption hits back to frames for visual answering, whereas dense caption RAG sends retrieved text without images.

\begin{table}[H]
\centering
\caption{Write--Read--Pack taxonomy. Main-search VLM invocations are capped at $K=40$ visual observations; cumulative harness cost may include multiple calls. Exploratory methods use different protocols.}
\label{tab:taxonomy}
\scriptsize
\begin{tabular}{p{0.17\textwidth}p{0.25\textwidth}p{0.28\textwidth}p{0.20\textwidth}}
\toprule
System & Write & Read & Pack \\
\midrule
\multicolumn{4}{l}{\emph{Reference constructors, $K=40$ per call}} \\
\uniformh & Retain the video stream; no separate index & Sample $K$ observations uniformly at answer time & Chronological \\
\clipknn & 320 frames and image embeddings & Question-to-image nearest neighbors & 40 frames, chronological \\
\captionknn & 320 frames, captions, and text embeddings & Caption nearest neighbors, then map hits back to their frames & 40 frames, chronological \\
Dense-caption text RAG & 320 captions and text embeddings; discard frames & Retrieve top-five captions as text; no images & Timestamped text list \\
\captionnav & 320 frames and captions; navigator reads at most 160 caption lines & VLM proposes one to four temporal ranges & Fill 40 frames within ranges \\
\midrule
\multicolumn{4}{l}{\emph{Main AKS-seeded search endpoint, $K=40$ per call}} \\
\cardinalityledger & Addressable video frames rendered into coarse and refined survey grids & Literal-interval bypass or L0$\rightarrow$L1 pointer refinement; counting queries additionally extract an ordered ledger & 40-frame local bursts plus global AKS anchors; optional text-only reconciliation \\
\midrule
\multicolumn{4}{l}{\emph{Exploratory: different capacity or non-promoted paths}} \\
Dense-caption pool & 640 captions and 320 display frames & Retrieve over 640 captions; display up to 320 frames & Segmented display with relevant markers \\
Motion-adaptive navigation & 320-frame probe, then motion reweighting; native uniform access only when $\leq320$ frames & Parent caption-navigation path & Fill 40 within proposed ranges \\
\bottomrule
\end{tabular}
\end{table}

\section{Search Archive and Cost Accounting}
\label{app:archive}
The archive contains the complete AKS-seeded trajectory, including executable source, parent identifiers, proposer traces, development scores, and raw-result pointers. We also report the complete ten-iteration uniform-seeded trajectory under the same strict frontier-promotion rule.

\begin{table}[H]
\centering
\caption{Complete ten-candidate AKS-seeded search. The parent changes only after a strict development-accuracy improvement; held-out results are not used during search.}
\label{tab:aks-main-search}
\scriptsize
\begin{tabular}{clrc}
\toprule
Candidate & Program & Dev. acc. & Frontier update \\
\midrule
0  & AKS parent               & 49.7 & --- \\
1  & ZoomRefine               & 47.4 & No \\
2  & StatedInterval           & 53.7 & Yes \\
3  & LegibleTimeline          & 52.6 & No \\
4  & OrdinalEnum              & 52.9 & No \\
5  & SurveyCommit             & 55.7 & Yes \\
6  & PointerLadder            & 57.4 & Yes \\
7  & CardinalityLedger        & \textbf{58.3} & Yes \\
8  & SectionedScan            & 58.0 & No \\
9  & ContiguousAnchors        & 58.3 & No (tie) \\
10 & SegmentTournament        & 54.9 & No \\
\bottomrule
\end{tabular}
\end{table}

\begin{table}[H]
\centering
\caption{Complete ten-iteration uniform-seeded VideoHarness-RSI trajectory. The development frontier changes only after a strict accuracy improvement. Held-out accuracy is shown only for the selected endpoint.}
\label{tab:uniform-search}
\small
\begin{tabular}{clrrc}
\toprule
Iteration & Program & Dev. acc. & Held-out acc. & Frontier update \\
\midrule
0  & Uniform-40                  & 36.0 & --- & --- \\
1  & CaptionNavigateVerify       & 43.4 & --- & Yes \\
2  & EmbedNavigate               & 48.6 & --- & Yes \\
3  & TemporalDirectionRetrieve   & 45.7 & --- & No \\
4  & ReasoningQueryBoost         & 46.6 & --- & No \\
5  & MMRRelevanceDiverse         & 48.0 & --- & No \\
6  & BurstRedecodeLocalize       & 42.6 & --- & No \\
7  & DualChannelAnswer           & 42.0 & --- & No \\
8  & MotionPairResolve           & 40.6 & --- & No \\
9  & \weakft                     & \textbf{50.9} & \textbf{49.0 (432/882)} & Yes \\
10 & AttributedPackAnswer        & 48.6 & --- & No \\
\bottomrule
\end{tabular}
\end{table}

The weak-seed trajectory yields three strict frontier updates. Search first moves beyond uniform sampling through navigation and embedding-based evidence retrieval, and later advances again to \weakft. The selected endpoint also improves over the uniform reference on held-out questions, while the final iteration does not replace the development frontier.

\section{CardinalityLedger Execution Analysis}
\label{app:cardinality-execution}

\begin{figure*}[t]
\centering
\includegraphics[width=0.96\textwidth]{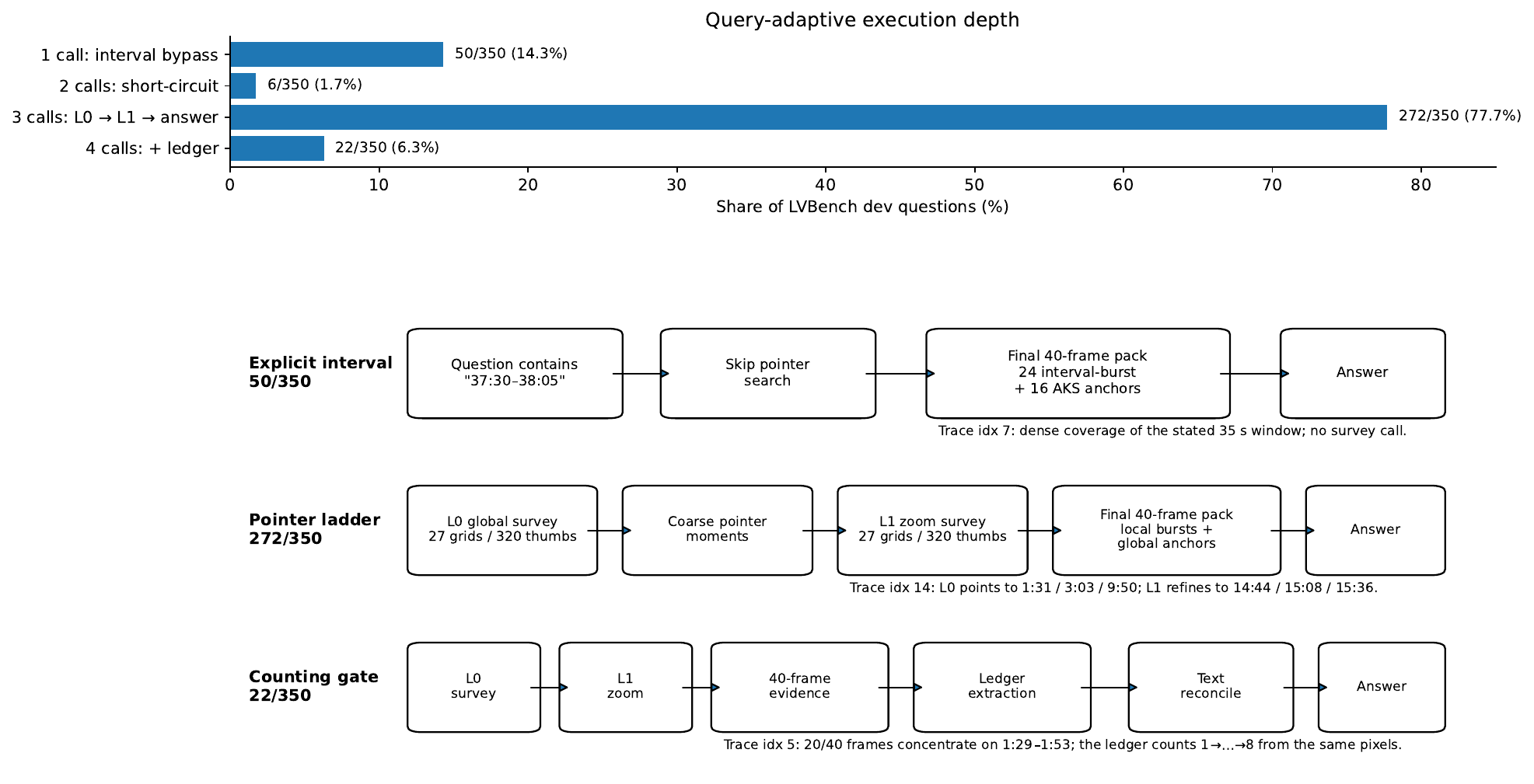}
\caption{Query-adaptive execution in \cardinalityledger, reconstructed from logged per-call frame provenance on LVBench development. Questions with a literal time interval bypass pointer search and directly combine a dense interval burst with global AKS anchors. The dominant path performs a coarse L0 survey, refines selected moments with an L1 survey, and then packs 40 answer frames from local bursts plus global AKS anchors. Counting questions additionally extract a ledger from the same visual evidence and reconcile it in text before answering. The bar plot reports the observed call-count distribution over all 350 development questions.}
\label{fig:cardinality-execution}
\end{figure*}

A representative explicit-interval query bypasses the surveys and concentrates answer frames inside the stated window. On the dominant ladder path, a long video is first summarized by coarse grids; pointer moments then define a narrower refinement stage, whose outputs determine local answer-time bursts while AKS anchors preserve global coverage. For counting questions, the harness reuses the selected visual evidence to construct an ordered ledger and performs a final text-only reconciliation. These traces show that RSI is discovering query-conditioned execution structure---when to bypass search, when to zoom, and when to add a symbolic verification step---rather than merely changing a frame-scoring function.

\section{Complementary Search Protocols}
\label{app:complementary}

\begin{table}[t]
\centering
\caption{Development-only visual-budget sweep. The oracle reads ground-truth evidence intervals and is non-deployable. No 882-question held-out result was logged for uniform $K=320$.}
\label{tab:budget}
\small
\begin{tabular}{rrrr}
\toprule
$K$ & Uniform & Oracle & Gap \\
\midrule
40 & 36.3 & 57.4 & 21.1 \\
80 & 38.9 & 58.6 & 19.7 \\
160 & 40.6 & 60.3 & 19.7 \\
320 & 46.9 & 55.7 & 8.9 \\
\bottomrule
\end{tabular}
\end{table}

\begin{figure}[H]
\centering
\includegraphics[width=0.8\linewidth]{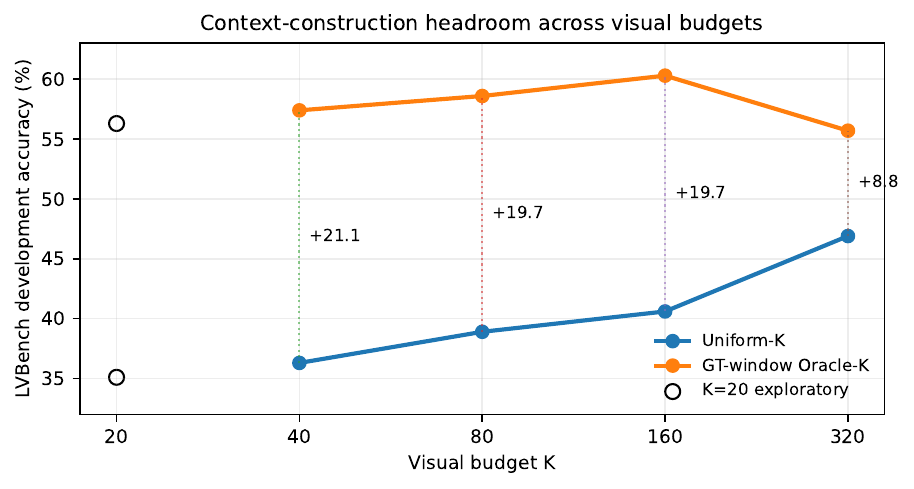}
\caption{Uniform sampling and a non-deployable evidence-window oracle across visual budgets on the 350-question development set. The remaining gap shows that context selection matters even as $K$ increases.}
\label{fig:budget-headroom}
\end{figure}

\begin{table}[t]
\centering
\caption{Exploratory high-capacity caption retrieval under a different 200/1,032 split. Paired counts are Uniform-only / Caption-only. The combined pool is descriptive and includes selected development examples.}
\label{tab:high-budget}
\scriptsize
\begin{tabular}{lrrrr}
\toprule
Split & Uniform & Caption ret. & $\Delta$ & McNemar \\
\midrule
Dev. 200 & 40.50 (81) & 44.50 (89) & +4.00 & 9/17; $p=.17$ \\
Test 1,032 & 42.83 (442) & 43.31 (447) & +0.48 & 61/66; $p=.72$ \\
Full 1,232 & 42.45 (523) & 43.51 (536) & +1.06 & -- \\
\bottomrule
\end{tabular}
\end{table}

The high-capacity caption-retrieval study uses a separate 200/1,032 split and is reported as a complementary capacity stress test. Its development-selected \prfh variant does not improve the matched uniform baseline on test.

\section{Literature-Inspired Harness Controls}
\label{app:literature-harnesses}
The WorldMM-style, Homer-style, and VideoSEAL-matched rows in Table~\ref{tab:main} map published memory and control-flow patterns into our common Write--Read--Pack interface, local LVBench subset, frozen Qwen answerer, and letter-only evaluation protocol. They serve as structural controls for how these design patterns behave under the same harness interface.

\paragraph{WorldMM-style.}
WorldMM builds complementary episodic, semantic, and visual memories and retrieves across modalities and temporal scales~\citep{vhYeo2026WorldMM}. Our controlled adaptation maps this pattern to event-oriented text, entity-oriented text, and timestamped visual stores queried by a VLM router before the common answer interface. \worldmmmax additionally forces visual-memory access and fills the available visual budget.

\paragraph{Homer-style.}
Homer combines keyframe, entity, and event memories with multi-round verification~\citep{vhJi2026Homer}. Our controlled adaptation retains this hierarchy inside a per-question constructor, retrieves from the three memories at answer time, and applies an answer-review step before returning the option.

\paragraph{VideoSEAL-matched.}
VideoSEAL separates long-horizon evidence seeking from answer authority through a planner--inspector architecture~\citep{vhQiu2026VideoSEAL}. Our controlled adaptation follows the same division: a planning stage retrieves candidate temporal spans from the local caption index, and a separate visual inspection stage receives the corresponding frames and produces the final answer. Its larger visual context is reported explicitly in Table~\ref{tab:main}.

\paragraph{Interpretation.}
Together, these rows test multimodal memory routing, hierarchical memory with verification, and decoupled planning and inspection under the same frozen-model harness interface.

\section{Reproducibility and Data Audit}
\label{app:repro}

The released artifact package fixes the seed-42 split, source-availability manifest, frozen-model settings, per-question predictions and correctness, and the complete reported ten-step trajectories for both initialization regimes. For the logged development-set executions of \aksh, \weakft, and \cardinalityledger, the released context sidecars preserve the ordered visual inputs, source-frame indices or timestamps, image hashes, request text, and visual-token accounting for each VLM invocation. These records support reconstruction and auditing of the visual context used by the reported development-time endpoint executions; equivalent packed-context logs are not claimed for the held-out evaluations.
\begin{table}[htbp]
\centering
\caption{Paired bootstrap confidence intervals on the 882-question LVBench held-out set. We use 20,000 paired resamples with seed 42 and percentile 95\% confidence intervals.}
\label{tab:bootstrap}
\small
\begin{tabular}{lrr}
\toprule
Comparison & $\Delta$ Acc. (pp) & 95\% CI (pp) \\
\midrule
\uniformh $\rightarrow$ \weakft & +12.7 & [9.4, 16.0] \\
\aksh $\rightarrow$ \cardinalityledger & +8.2 & [5.1, 11.2] \\
AKS-90 $\rightarrow$ \cardinalityledger & +5.6 & [2.4, 8.7] \\
\bottomrule
\end{tabular}
\end{table}

On the 882-question held-out set, the aligned AKS--\cardinalityledger comparison has 63 AKS-only and 135 \cardinalityledger-only correct predictions, yielding an exact two-sided McNemar test of $p=3.39\times10^{-7}$. We additionally use paired bootstrap resampling with 20,000 resamples, seed 42, and percentile 95\% confidence intervals. Table~\ref{tab:bootstrap} reports these intervals. The release also includes scripts that verify the reported main scores and paired McNemar statistics directly from the stored dumps. Because outer-loop proposals are generated by a proprietary LLM service, the archive supports auditing the reported search trajectories but does not guarantee deterministic regeneration of the same proposal sequence.

\bibliographystyle{plainnat}

\clearpage
\section*{Responsible Use Statement}

This work studies automated search over executable context-construction harnesses for long-video understanding. By improving how a frozen vision-language model acquires and organizes evidence from long videos, such techniques may increase the efficiency and adaptability of video-analysis systems. These capabilities may be beneficial in applications such as video retrieval, accessibility, education, and large-scale media analysis, but they may also be used in privacy-sensitive settings, including monitoring or surveillance, where more effective automated video understanding could amplify existing concerns around consent, privacy, and misuse.

Automated recursive harness optimization also introduces technical risks. Search procedures may overfit development feedback, exploit benchmark-specific regularities, discover brittle or difficult-to-audit execution strategies, or increase inference cost through additional model calls and context construction. Improvements on benchmark accuracy therefore do not by themselves guarantee reliability, robustness, or suitability for deployment in high-stakes environments.

We mitigate these risks in the present study by restricting experiments to benchmark evaluation, freezing the downstream vision-language model and answer interface, enforcing a bounded visual context for each model invocation, and separating development feedback from held-out and cross-benchmark evaluation. We retain released endpoint context logs and archived search artifacts to improve auditability, and report cumulative inference cost alongside accuracy so that gains are not considered independently of resource usage.

Any deployment of automated harness-search techniques in privacy-sensitive, safety-critical, or high-impact settings should involve additional safeguards beyond those evaluated here. These may include appropriate data governance and consent procedures, human oversight, task-specific robustness and safety evaluation, access controls, logging and auditing of execution behavior, and continuous monitoring for distribution shift or unintended use. We do not advocate deploying the systems studied in this work without such application-specific evaluation and governance.

\end{document}